\documentclass[10pt,twocolumn,letterpaper]{article}

\usepackage{iccv}
\usepackage{times}
\usepackage{epsfig}
\usepackage{graphicx}
\usepackage{amsmath}
\usepackage{amssymb}

\usepackage[breaklinks=true,bookmarks=false]{hyperref}

\iccvfinalcopy % *** Uncomment this line for the final submission

\def\iccvPaperID{****} % *** Enter the ICCV Paper ID here

\ificcvfinal\fi

\begin{document}

%%%%%%%%% TITLE
\title{Neighborhood Consensus Contrastive Learning for Backward-Compatible Representation}

\author{
{Shengsen Wu{$^{1,2,*}$}, Liang Chen{$^{3,*}$}, Yihang Lou{$^{4}$}, Yan Bai{$^{2}$}, Tao Bai{$^{4}$}, Minghua Deng{$^{3}$} and Lingyu Duan{$^{2}$}} \\
  {$^{1}$} The SECE of Shenzhen Graduate School, Peking University \\
  {$^{2}$} Institute of Digital Media, Peking University \\
  {$^{3}$} School of Mathematical Sciences, Peking University \\
  {$^{4}$} GoTen AI Lab, Intelligent Vision Dept, Huawei Technologies\\
  %\thanks{This work is completed in Huawei Co.Ltd. * means the equal contribution.} 

}

\maketitle
% Remove page # from the first page of camera-ready.
\ificcvfinal\thispagestyle{empty}\fi
  \let\thefootnote\relax\footnotetext{This work is completed in Huawei Co.Ltd. * means the equal contribution.}

\begin{abstract}
In object re-identification (ReID), the development of deep learning techniques often involves model updates and deployment. It is unbearable to re-embedding and re-index with the system suspended when deploying new models. Therefore, backward-compatible representation is proposed to enable ``new'' features to be compared with ``old'' features directly, which means that the database is active when there are both ``new'' and ``old'' features in it. Thus we can scroll-refresh the database or even do nothing on the database to update. % even do not need to update the database.

The existing backward-compatible methods either require a strong overlap between old and new training data or simply conduct constraints at the instance level. Thus they are difficult in handling complicated cluster structures and are limited in eliminating the impact of outliers in old embeddings, resulting in a risk of damaging the discriminative capability of new features.  In this work, we propose a Neighborhood Consensus Contrastive Learning (NCCL) method. With no assumptions about the new training data, we estimate the sub-cluster structures of old embeddings. A new embedding is constrained with multiple old embeddings in both embedding space and discrimination space at the sub-class level. The effect of outliers diminished, as the multiple samples serve as ``mean teachers''.
Besides, we also propose a scheme to filter the old embeddings with low credibility, further improving the compatibility robustness. Our method ensures backward compatibility without impairing the accuracy of the new model. And it can even improve the new model's accuracy in most scenarios. % 可以写strong overlap 吗，确认哈，我英语一般

%With no assumptions about the new training data, we conduct constraints with multiple positives and negatives at the sub-class level, thus the effect of outliers in old embeddings diminished as multiple old embeddings serve as ''mean teachers”. % 是否会有歧义   上下文衔接  下面那句太具体了
%The sub-class view is estimated by the similarity between old features, and different sub-clusters have different priorities. Such constraints work in both embedding structures and discriminative knowledge. 
%Besides, we also propose a scheme to filter the old embeddings with low credibility, which can further improve the compatibility robustness. Our method ensures backward compatibility without impairing the accuracy of the new model. And it can even improve the new model's accuracy in most scenarios.

\end{abstract}

\section{Introduction} %已重写
General object re-identification (ReID) aims at finding a significant person/vehicle of interest in a large amount of collection, called ``gallery''.
It has attracted much interest in computer vision due to its great contributions in urban surveillance and intelligent transportation~\cite{Ye2021, Khan2019}. Specifically, existing ReID methods typically map each image into vector space using a convolutional neural network. 
The output in vector space is usually called ``embedding''.
Given query images of a specific object, the process of ReID is to find the same identity in the gallery by finding its nearest neighbors in embedding space.
%%这个怎么突出identify

\begin{figure}
  \centering
  \centerline{\includegraphics[width=1\columnwidth]{"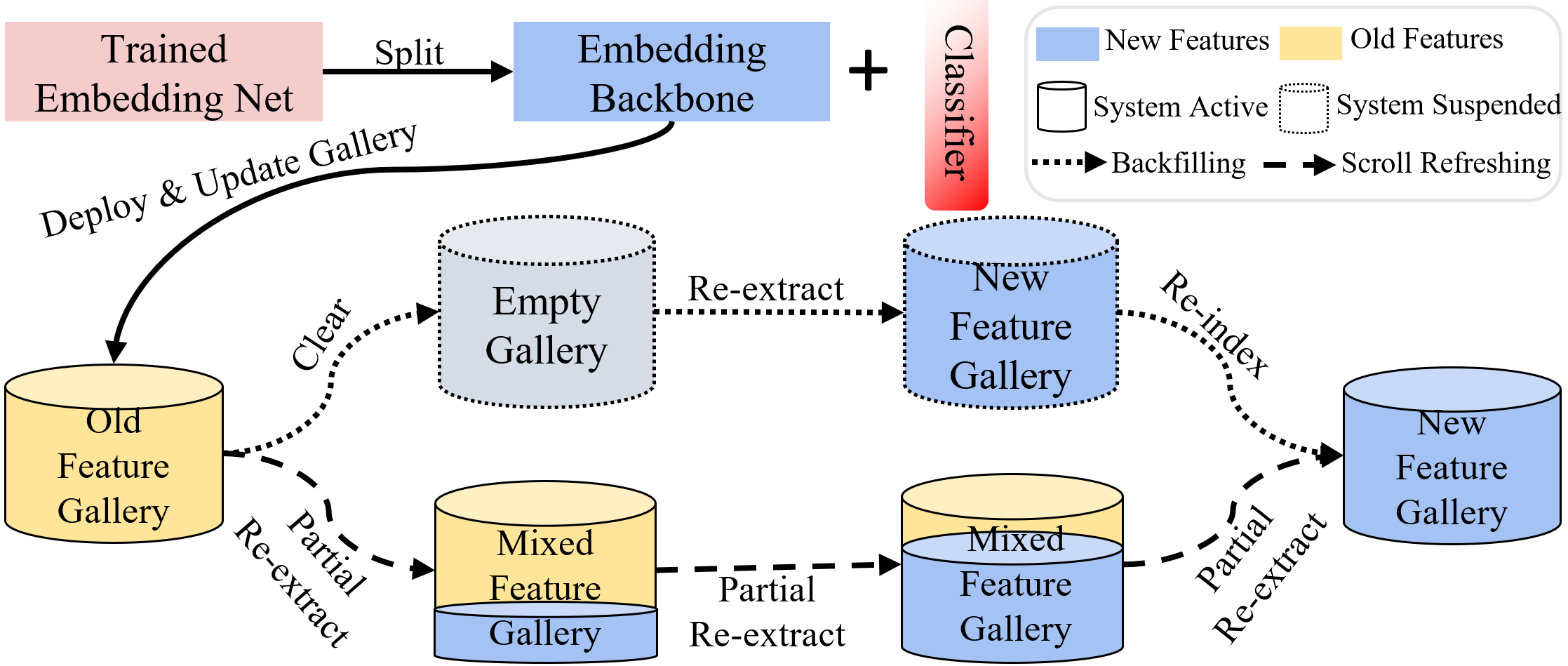"}}   % 把gallery和特征联系在一起  activate/inactivate gallery 避免   re-extract/replace
  \caption{The Deployment Process. Without backward compatibility, the process to update the gallery is ``backfilling'' with the system suspended until re-index has been done. With backward compatibility, the process to update is ``scroll refreshing'' with the system always active. Notice that the classifier is abandoned when deployed as it is useless for retrieval but only a waste of space. Thus the old classifier is unavailable when the old model is from a third party. }
  \label{deploy}
\vspace{-0.6cm} 
\end{figure}

As the new training data or the improved model designs are available, the deployed model will be updated to a new version with better performance. Then we need to refresh the gallery to harvest the benefits of the new model, namely ``backfilling''. As it is common to have millions or even billions of images in the database/gallery \cite{Radenovic2018}, backfilling is a painful process, which requires re-extracting embeddings and re-building indexes with the system suspended, as shown in Figure \ref{deploy}. An ideal solution is enabling ``new'' embeddings to be compared with ``old'' embeddings directly. Then we can gradually replace ``old'' embedding with ``new'' embeddings in the gallery while the system is still active, named ``scroll refreshing''. What's more, if the gallery is in scroll-refreshed mode initially (\textit{e.g.}, only keep the surveillance data in the past $x$ days), the proportion of new embeddings in the gallery increases from 0\% to 100\% over some time automatically. % after the deployed model is updated. 

To achieve backfilling-free, backward-compatible representation learning methods have recently arisen. 
Under a backward-compatible setting, given an anchor sample from new embeddings, no matter the positives and negatives from old/new embeddings, the anchor is consistently closer to the positives than the negatives. Recently, metric learning methods are adopted to optimize the distances between new and old embeddings in an asymmetric way \cite{Budnik2020}. 
However, the adopted pair-based triplet losses can not well consider the feature cluster structure differences across different embeddings. They are sensitive to the outliers in the old embeddings and hard to convergence on the large-scale dataset. 
Alternatively, the classifier of the old model is used to regularize the new embeddings in training the new model \cite{Shen2020}, termed as ``backward-compatible training (BCT)''. 
The old classifier is based on a weight matrix, where each column can be viewd as a ``prototype''  of a particular class in old training sets. Such prototype is also limited in characterizing the cluster structure.
What's more, BCT can only work when the new and old training sets have common classes and the old classifier is available.

The old embedding space is often not ideal and faces complicated cluster structures. 
For example, due to the intra-class variance caused by varied postures, colors, viewpoints, and lighting conditions, the cluster structure often tends to be scattered as multiple sub-clusters. 
Moreover, the old embedding space also inevitably contains some noise data and outliers in clusters. 
Simply constraining the learned ``anchor'' of new embeddings close to the randomly sampled ``fixed positives'' of old embeddings has a risk of damaging the discriminative capability of new embeddings. 
Therefore, it is necessary to consider the entire cluster structure across new and old embeddings in backward-compatible learning.

% 重写 和abstract 改一下 抽象     poor features  constrbutions little when optimizing.    主导梯度的主要贡献  dom  the grad.. contribution.   isdiminished -> 缓解

%In this work, we propose a Neighborhood Consensus Contrastive Learning (NCCL) method for backward-compatible representation. Previous methods mainly constrain the distributions of the new embeddings to be consistent with the old. We design a supervised contrastive learning method to constrain the distance relationship between the new and old embeddings to be consistent. Specifically, a new embedding is supervised by multiple old embeddings at a cluster view. Multiple old embeddings serve as ``mean teacher'', thus the gradient corresponding to poor old embedding is diminished and averaged when backpropagation. The cluster view is estimated by the similarity between old embeddings. Old embeddings in different clusters have different priorities when optimizing.   

In this work, we propose a Neighborhood Consensus Contrastive Learning (NCCL) method for backward-compatible representation. Previous methods mainly constrain the distributions of the new embeddings to be consistent with the old. We design a supervised contrastive learning method to constrain the distance relationship between the new and old embeddings to be consistent. Specifically, we estimate the sub-cluster structures in old embeddings.  A new embedding is constrained by multiple old embeddings from different sub-clusters. The priorities of different sub-clusters are carefully designed to dominate the optimizing progress.
Also, the effect of outliers in old embeddings diminished, as the multiple old embeddings serve as ``mean teachers” and the outliers contribute little during the optimizing progress. 
Typically, the classifier head in a network converts the embedding space to discrimination space for classification. Therefore, the discrimination space contains rich class-level information \cite{zhong2020bi}. Correspondingly, we further design a soft multi-label task to exploit such knowledge to enhance compatible learning. Such labels are an auxiliary apart from the real-word labels, making the new embeddings more discriminative. Besides, since the quality of the old model is unpredictable, we estimate the entropy distribution of the old embeddings and remove the ones with low credibility. It helps to maintain robustness when learning backward-compatible representation from old models with various qualities.

Our contributions are summarized as follows:
    \vspace{-0.3em}

\begin{itemize}       

\vspace{-0.3em}
    \item 
    %We propose a Neighborhood Consensus Contrastive Learning (NCCL) method for backward-compatible representation. The NCCL constraints the new embeddings with multiple old embeddings at the sub-class level. The effect of poor old embeddings is  %The NCCL exploits the neighborhood structure in both the embedding and discrimination spaces.
    %强调思想   邻域结构   不应该全落在sub-clusters,  neighbour 更好地消除 outliers , compatible 
    %Unlike the existing methods that directly adopt metric learning methods asymmetrically, w
     %We propose a Neighborhood Consensus Contrastive Learning (NCCL) method for backward-compatible representation. With no assumptions about the new training data, we start from a neighborhood consensus perspective, constrain at the sub-cluster level to make the optimization process smoothly, and manage to reduce the contribution of outliers in such an optimization process. 
     
     %We propose a Neighborhood Consensus Contrastive Learning (NCCL) method for backward-compatible representation. With no assumptions about the new training data, we constrain the new embeddings with multiple old embeddings at the sub-class level. The effect of outliers diminished, and the neighborhood structure is estimated to keep the discriminative capability of new embeddings.
     We propose a Neighborhood Consensus Contrastive Learning (NCCL) method for backward-compatible representation. With no assumptions about the new training data, we start from a neighborhood consensus perspective, and constrain at the sub-cluster level to keep the discriminative capability of new embeddings. The impact of outliers is reduced as they contribute little to such optimizing progress.%``Mean teachers'' is adopted to reduce the impact of outliers. 

     %We propose a Neighborhood Consensus Contrastive Learning (NCCL) method for backward-compatible representation. We start from a neighborhood consensus perspective, and constrain at the sub-cluster level to make the optimization process smoothly. The proposed NCCL can reduce the impact of the outliers and is not limited to any assumption of the new training data.
     
     % We propose a Neighborhood Consensus Contrastive Learning (NCCL) method for backward-compatible representation. We start from a neighborhood consensus perspective, and constrain at the sub-cluster level to make the optimization process smoothly. NCCL can reduce the impact of the outliers and is not limited to any assumption of the new training data.
     
    %We consider the particularity of backward-compatible learning: the sub-cluster structures of the fixed old embeddings make it difficult to optimize; the outliers in the fixed old embeddings may impact the new embeddings. Therefore, we start from a neighborhood consensus perspective, constrain at the sub-cluster level to make the optimization process smoothly, and manage to reduce the contribution of outliers in such an optimization process.    %Different from existing methods which adopt metric learning method in asymmetric way directly, we have considered the optimization difficulties caused by the old model is fixed and there are sub-clusters and outliers in the process of learning backward-compatiaby.
     \vspace{-0.3em}

    %We propose a Neighborhood Consensus Contrastive Learning (NCCL) method for backward-compatible representation. With no assumptions about the new training data, we constrain the new embeddings with multiple old embeddings at the sub-class level. The effect of outliers diminished, and the neighborhood structure is estimated to keep the discriminative capability of new embeddings.
   %太单薄了 第二点扩展一下   
   \item 
   We perform backward-compatible learning from both embedding space and discrimination space, with priorities of different sub-clusters dominating the optimizing progress. We further propose a novel scheme to filter the old embeddings with low credibility, which is helpful to maintain robustness with the qualities of old models various.%against the noise data in old models. 
       \vspace{-0.3em}

    \item The proposed method obtains state-of-the-art performance on the evaluated benchmark datasets. We can ensure backward compatibility without impairing the accuracy of the new model. In most cases, it even improves the accuracy of the new model.%, \textit{e.g.} ResNet50+MSMT17 (mAP +7.55\%), ResNet50+VeRi-776 (mAP +4.72\% ).%, especially on the MSMT17 (mAP +7.55\%) and VeRi-776 (mAP +4.72\%).
\end{itemize}

\section{Related work} % 待重写
\subsection{Object re-identification}
%Object re-identification technology is widely used in target tracking and intelligent security, mainly including person re-identification and vehicle re-identification. It can realize cross-camera image retrieval, tracking, and trajectory prediction of pedestrians or vehicles. In recent years, driven by deep learning technology, research in the object re-identification field has mainly focused on network architecture design \cite{Wang2018, ZhouK2019, Quan2019}, feature representation learning \cite{Zhao2017, Yao2019}, deep metric learning \cite{Wojke2018, Chen2018, Guo2018, Zhong2019}, ranking optimization \cite{Ye2015, Bai2019}, and recognition under video sequences \cite{Hou2019, Fu2019, Li2019}. This paper is to learn backward-compatible representation for ReID models. 
Object re-identification can realize cross-camera image retrieval, tracking, and trajectory prediction of pedestrians or vehicles. In recent years, research in this field has mainly focused on network architecture design \cite{Wang2018, ZhouK2019}, feature representation learning \cite{Zhao2017, Yao2019}, deep metric learning \cite{Wojke2018, Chen2018, Zhong2019}, ranking optimization \cite{Ye2015, Bai2019}, and recognition under video sequences \cite{Hou2019, Fu2019, Li2019}. 
Deep metric learning aims to establish similarity/dissimilarity between images.
This paper is to design based on the metric learning paradigms and establish similarity/dissimilarity between new and old embeddings, aiming to learn backward-compatible representation.  %这个和后面倒过来

\subsection{Backward-compatible representation}
Backward-compatible learning encourages the new embeddings closer to the old embeddings with the same class ID than that with different class IDs. It is an emerging topic. To the best of our knowledge, there only exist two pieces of research.
Budnik \emph{et al.} \cite{Budnik2020} adopts the metric learning loss,  \textit{e.g.}, triplet loss and contrastive loss, in an asymmetric way to achieve backward-compatible learning. We call this method “Asymmetric” for convenience. 
Shen \emph{et al.} \cite{Shen2020} proposes BCT, a backward-compatible representation learning model. BCT feeds the new embeddings into the old classifier. Then the output logits are optimized with cross-entropy loss. 
BCT and Asymmetric are essentially identical, as it is equivalent to a smoothed triplet loss where each old embedding class has a single center \cite{qian2019softtriple}. With the newly added data, BCT needs to use knowledge distillation techniques. In general, existing methods simply adopt embedding losses or classification losses in an asymmetric way. However, such loss is limited in handling the complicated cluster structures of fixed old embeddings. In this article, we consider this problem from a neighborhood consensus perspective with both embedding structure and discrimination knowledge.  %discrimination knowledge
%In general, existing methods simply conduct constraints in the embedding space or discriminative space and ignore the intra-class variance of the old embeddings. In this article, we consider this problem from a neighborhood consensus perspective with both embedding structure and discriminative knowledge.

\subsection{Contrastive learning} %这段还是不太好 我再想想

Contrastive learning is a sub-topic of metric learning, which learns representations by contrasting positive pairs against negative pairs. Asymmetric has adopted a contrastive loss \cite{2006Dimensionality} in backward-compatible learning. However, it is margin-based and only able to capture local relationships. Another kind of contrastive learning paradigm is based on InfoNCE \cite{oord2018representation}, which proves its contrastive power increases with more positives/negatives. It has been widely used in unsupervised representation learning tasks with various variants, including instance-based \cite{ChenX2020, ChenT2020b, Grill2020}, cluster-based \cite{Caron2020, LiJ2020}, mixed instance and cluster-based \cite{wang2020unsupervised, Zhong2020}. In this article, we start from InfoNCE and design a compatible learning framework. Specifically, we generalize InfoNCE to constrain the new embeddings in both embedding space and discrimination at the sub-class level. %在不同层面上 落脚点小

\section{Methods}
\label{methods}
\subsection{Problem formulation} %已重写
Following the formulation in \cite{Shen2020}, a embedding model contains two modules: an embedding function $\phi: \mathcal{X}\rightarrow \mathcal{Z}$ that maps any query image $x\in \mathcal{X}$ into vector space $\mathcal{Z}$ and a classifier $\varphi: \mathcal{Z}\rightarrow \mathcal{R}^{K}$. 
Assume that we have an old embedding function $\phi_{old}$ trained on the old training dataset $\mathcal{D}_{old}$ with $K_{old}$ IDs. Then we have a new embedding function $\phi_{new}$ and a new classifier $\varphi_{new}$ trained on the new training data $\mathcal{D}_{new}$ with $K_{new}$ IDs. $\mathcal{D}_{new}$ can be a superset of $\mathcal{D}_{old}$. The label space is set to $\mathcal{Y}$.

% We define the gallery set as $\mathcal{G}$, and the embeddings of images in $\mathcal{G}$ are obtained by $\phi_{old}$. The query images are transformed to embeddings by $\phi_{old}$ in the test stage. Assume that $M$ is a metric, e.g., mAP or top-$k$ accuracy, to evaluate the effectiveness of the retrieval between the gallery set and the query set, then $M(\phi_{old}, \phi_{old})$ is the self-test result using the old embedding model. Once the system is updated, similarly, $M(\phi_{new}, \phi_{new})$ is the self-test result based on the $\phi_{new}$ for both gallery and query. Since the number of images in $\mathcal{G}$ could be extremely large, it is costly to refresh the embeddings in $\mathcal{G}$. Therefore, $M(\phi_{new}, \phi_{old})$ refers to the cross-test result using $\phi_{new}$ for query and $\phi_{old}$ for the gallery. Empirically, the goal of backward-compatible learning is to accomplish $M(\phi_{new}, \phi_{new}) \geq M(\phi_{new}, \phi_{old}) \geq M(\phi_{old}, \phi_{old})$.

We define the gallery set as $\mathcal{G}$, and query set as $\mathcal{Q}$. $M(\phi_{m_1},\phi_{m_2})$, $\forall(m_1,m_2)$$\in$$\{new,old\}$ is an evaluation metric, 
\textit{e.g.}, mAP, to evaluate the effectiveness of the retrieval when $\mathcal{Q}$ is processed by $\phi_{m_1}$ and $G$ processed by $\phi_{m_2}$. 
$M(\phi_{new}, \phi_{new})$ $\geq$  $M(\phi_{old}, \phi_{old})$ is naturally established, as it is why we update the model. 
% $M(\phi_{new}, \phi_{old})$  说得清楚一点，还是有点抽象
When $\phi_{new}$ is not compatiable with $\phi_{old}$, $M(\phi_{new}, \phi_{old})$ is very low, even 0\%, which means retrievals in the ``old gallery'' with ``new query'' return with poor results. We have to suspend the system and backfill $\mathcal{G}$ to harvest $M(\phi_{new}, \phi_{new})$. However, when $\phi_{new}$ is compatiable with $\phi_{old}$, $M(\phi_{new},  \phi_{old})$ is comparable than $M(\phi_{old}, \phi_{old})$ or even outperforms it. Thus $\mathcal{G}$ can be scroll-refreshed with the system active. Empirically, the goal of backward-compatible learning is to accomplish $ M(\phi_{new}, \phi_{old}) \geq M(\phi_{old}, \phi_{old})$ without impacting $M(\phi_{new}, \phi_{new})$.

\subsection{Backward-compatible criterion}
\begin{figure}
  \centering
  \centerline{\includegraphics[width=1\columnwidth]{"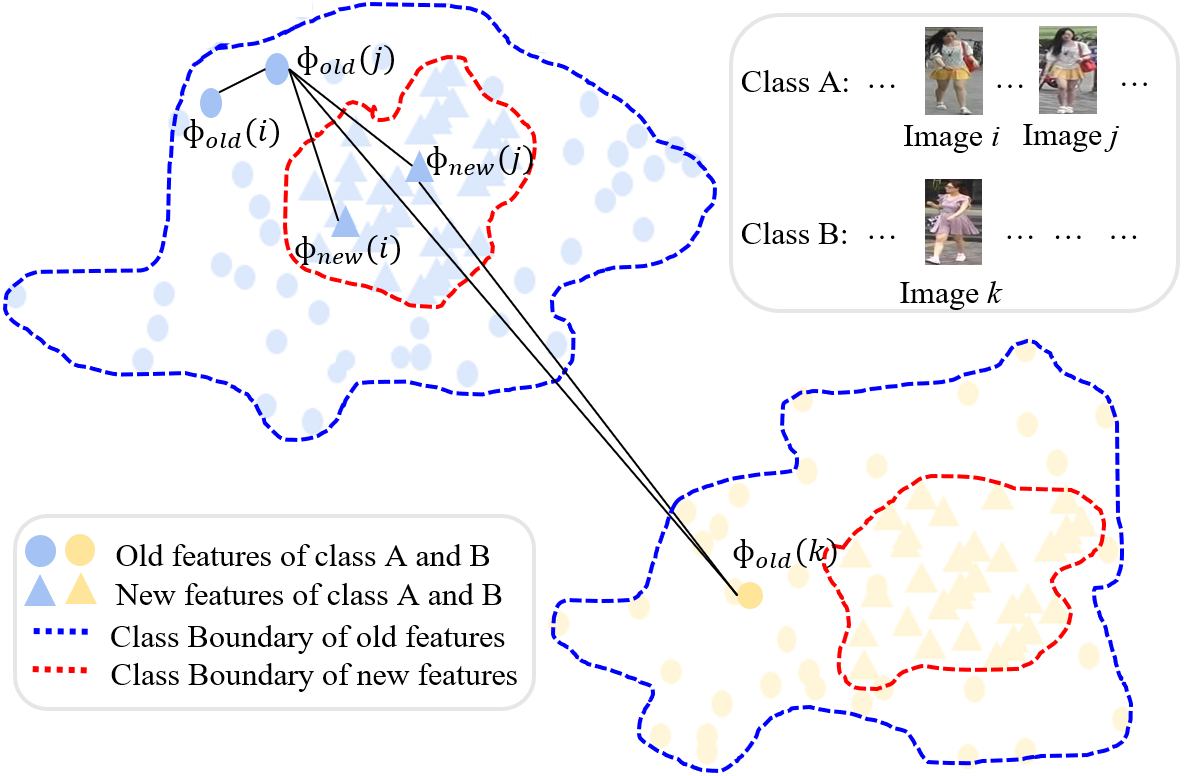"}}
  %\centerline{\epsfig{figure="./Figs/criterion.png",width=8cm}}
  %\caption{New/old embeddings in the vector space. We can find a triplet ($A$, $B$, $C$) that  $d(\phi_{new}(B), \phi_{old}(C) \textless d(\phi_{old}(B), \phi_{old}(C)$ and $d(\phi_{new}(A), \phi_{old}(B) \textgreater d(\phi_{old}(A), \phi_{old}(B)$. }%待补充
  \caption{The distance relationship in the feature space. $\phi_{new}$ is compatiable with $\phi_{old}$, but pair $(i,j)$ and $(j,k)$ does not satify Inequality (\ref{equation 1}) and (\ref{equation 2}). }
  \label{fig1}
  \vspace{-0.5cm} 

\end{figure}

%The constraints of backward-compatible compatibility aim at accomplishing $M(\phi_{new}, \phi_{old})$ $\geq$  $M(\phi_{old}, \phi_{old})$ without impacting $M(\phi_{new}, \phi_{new})$.

Shen \emph{et al.} \cite{Shen2020} give a strict criterion of backward-compatible compatibility in BCT. 

  \vspace{-0.6cm} 
\begin{align}
d(\phi_{new}(i), \phi_{old}(j)) & \geq d(\phi_{old}(i), \phi_{old}(j)), \nonumber \\
& \forall (i,j)   \in \{(i,j): y_{i}\neq y_{j}\}. 
\label{equation 1}
\end{align}
  \vspace{-0.7cm} 
\begin{align}
d(\phi_{new}(i), \phi_{old}(j) & \leq d(\phi_{old}(i), \phi_{old}(j)), \nonumber \\
& \forall (i,j) \in \{(i,j): y_{i}=y_{j}\}. 
\label{equation 2}
\end{align}
  \vspace{-0.5cm} 
  
%Such definition may lead to the increase of variance in the new embedding classes in some cases. As shown in Figure \ref{fig1}, the new embeddings $A_{new}$ and $B_{new}$ are obviously backward-compatible with their old embeddings. 
%However, we can find $d(B_{new}, C_{old}) \textless d(B_{old},C_{old})$ and
%d(A_{new}, B_{old}) \textgreater d(A_{old}, B_{old})$, which do not satisfy Eq. \ref{equation 1} and \ref{equation 2}.
%that the distance relationship between $B_{old}$, $B_{new}$ and $C_{old}$ deviates from the inequality (\ref{equation 1}), and the distance relationship between $A_{old}$, $A_{new}$ and $B_{old}$ deviates from the inequality (\ref{equation 2}). 
%Constraining $A_{new}$ and $B_{new}$ to satisfy the inequality will lead to sub-optimal solutions.
%Such definition may lead to the increase of variance in the new embedding classes in some cases. As shown in Figure \ref{fig1}, $\phi_{new}$ is obviously backward-compatible with $\phi_{old}$. However, we can find $d(\phi_{new}(j),\phi_{old}(k))$ $\le$$ d(\phi_{old}(j),\phi_{old}(k))$, $d(\phi_{new}(i),\phi_{old}(j))$$ \ge$$ d(\phi_{old}(i),\phi_{old}(j)) $, which do not satisfy Eq. (\ref{equation 1}) and (\ref{equation 2}).

  \noindent
Such definition is the sufficient and unnecessary condition to achieve compatiblity, and it may impact $M(\phi_{new}, \phi_{new})$. As shown in Figure \ref{fig1}, $\phi_{new}$ is compatible with $\phi_{old}$. However, we can find $d(\phi_{new}(j)$$,\phi_{old}(k))$$ \le$$ d(\phi_{old}(j)$$,\phi_{old}(k))$, $d(\phi_{new}(i)$$,\phi_{old}(j))$$ \ge$$ d(\phi_{old}(i)$$,\phi_{old}(j)) $, which do not satisfy Inequality (\ref{equation 1}) and (\ref{equation 2}). Constraining $\phi_{new}$ to satisfy them enlarge the class boundary of new features. Therefore, we come up with a new backward-compatible criterion:  
%second, in image retrieval task, for each query image, it requires the neighboring image to maintain the same ID tag, and the images far away from it represent other IDs. Therefore, we give our backward-compatible compatible criterion below,
  \vspace{-0.2cm} 
\begin{align}  
d(\phi_{new}(i), \phi_{old}(j)) < d(\phi_{new}(i), \phi_{old}(k)), \nonumber \\
\forall (i,j,k)\in \{(i,j,k): y_{i}=y_{j}\neq y_{k}\}.
\label{equation 3}
\end{align}
  \vspace{-0.5cm}

    %\vspace{-0.3cm} 
   \noindent
Inequality (\ref{equation 3}) is the minimal constraints to accomplish $M(\phi_{new}, \phi_{old})$ $\geq$  $M(\phi_{old}, \phi_{old})$. 
%, precisely describes the relationship between the new and old models when they are backward-compatible. 
Given image $i$, the distance between anchor (termed as $\phi_{new}(i)$) and other fixed positives (termed as $\phi_{old}(j)$) is smaller than the distance between it and fixed negatives (termed as $\phi_{old}(k)$). 
%Given a image $x_{i}$, the distance between the new embedding $\phi_{new}(x_{i})$(``anchor'') and other old embeddings $\phi_{old}(x_{j})$ with the same ID(``fixed positives'') is smaller than the distance between it and different ID samples' old features $\phi_{old}(x_{k})$.

\subsection{Neighborhood consensus compatible learning}
\begin{figure*}
  \centering
  \centerline{\includegraphics[width=2.15\columnwidth]{"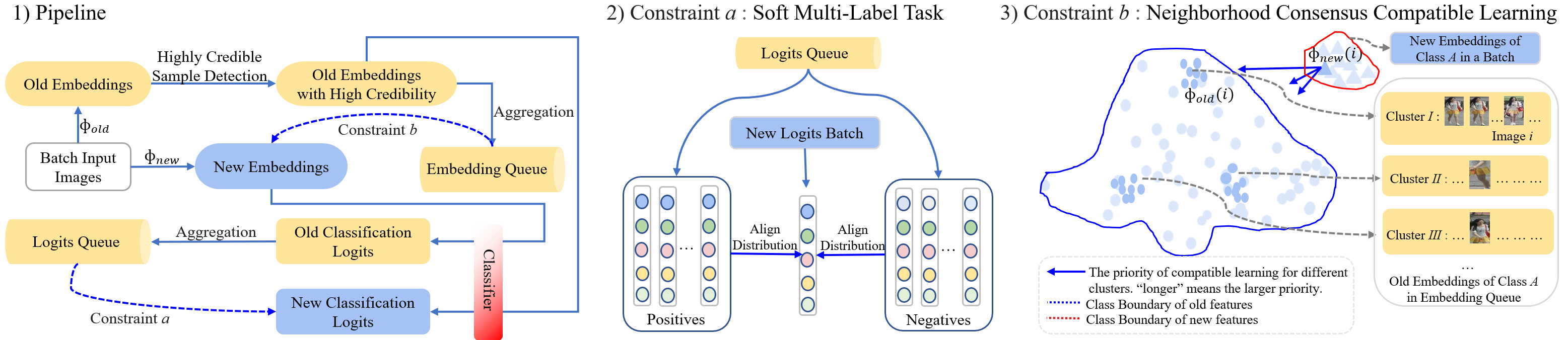"}}
  %\centerline{\epsfig{figure="./Figs/method.jpg",width=17.5cm}}
  %\caption{An illustration of our method NCCL. The left part shows the highly credible sample selection and contrastive learning in the embedding space; The middle part shows that each anchor gives different attention weights for different positives; The right part shows that the pseudo-multi-labels are used to align the discrimination distribution.} 
  \caption{An illustration of our method NCCL. The left part shows the pipeline of using the old embeddings to regularize the new embeddings; The middle part shows that we use a soft multi-label task to align the discrimination distribution; The right part shows that we learn the compatible embeddings at cluster view. } 
  \label{fig:pipeline}
  %\vspace{10em}
    %\vspace{-0.3cm} 

\end{figure*}
%\vspace{-10pt}

To satisfy Inequality (\ref{equation 3}), a naive solution is to adopt the metric learning loss in an asymmetric way, \textit{e.g.}, triplet loss \cite{Budnik2020} and cross-entropy \cite{Shen2020} 
. However, there are several limitations: 1) The potential manifold structure of the data can not be characterized by random sampling instances or classifiers \cite{qian2019softtriple}. 2) They are susceptible to those boundary points or outliers with the newly added data, making the training results unpredictable. 3) They treat all the old embeddings equally important, which can be overwhelmed by less informative pairs, resulting in inferior performance \cite{wang2019multi}. 

%In this work, we propose a neighborhood consensus supervised contrastive loss with multiple positives and negatives to exploit the neighborhood structural information. The InfoNCE based contrastive loss\cite{oord2018representation} is commonly used for self-supervised learning, proving that its contrastive power increases with more positives and negatives. To flexibly select neighborhoods, a FIFO memory queue is adapted to store the fixed positives/negatives which are updated by batches, termed as $\mathcal{B}$. It can represent the potential manifold structural information of the old embedding space.
In this work, we propose a neighborhood consensus supervised contrastive learning method for backward-compatible learning. We proposed the neighborhood consensus weight to estimate the potential manifold structure. 
A FIFO memory queue is adapted to store the estimated clusters, updating by batches, termed as $\mathcal{B}$. Then we propose a contrastive loss to regularize the new embeddings with the old clusters. 
Define $A(i)\equiv \mathcal{B} \diagdown \{i\} $ and $P(i)\equiv \{p\in A(i): y_{p}=y_{i}\}$, our training objective on the embedding space is a weighted generalization of contrastive loss, given  by
  \vspace{-0.1cm} 
\begin{align}
\label{equation 4} L_{1}  &= \sum_{i\in \mathcal{D}_{new}}\sum_{p\in P(i)} -w_{ip}\log s_{ip}, \\
\label{equation 5} w_{ip} &= \frac{1}{2}(\frac{\phi_{old}(i)\phi_{old}(p)}{||\phi_{old}(i)||||\phi_{old}(p)||} + 1) \in [0, 1], \\
\label{equation 6} s_{ip}  &=\frac{\exp(\phi_{new}(i)\phi_{old}(p)/\tau)}{\sum_{a\in A(i)}\exp(\phi_{new}(i)\phi_{old}(a)/\tau)}. 
%\label{equation 4}
\end{align}
%, where $\tau$ is a temperature parameter that controls the sharpness of similarity distribution.
  \vspace{-0.3cm}

$w_{ip}$ estimates the sub-cluster structures by measuring the affinity relationship between anchors $i$ and fixed positives $p$.  
Ideally, each new anchor embedding should focus on the fixed positives from its corresponding old embedding cluster.
% blurred 确认一下
However, the boundaries between local clusters may be blurred. Thus we adopt the soft weight to concentrate more on the positives, which are more informative. ``More informative'' means a higher probability to belong to a local cluster corresponding to the anchor. 
Given a positive image pair $(i,p)$ and view $i$ as the anchor, it is intuitive that the weight $w_{ip}$ of $\phi_{old}(p)$ rises in inverse proportion to the distance between $\phi_{old}(i)$ and $\phi_{old}(p)$.
Specifically, $w_{ip}$ is calculated on the embeddings after L2 normalized by the cosine distance, regarded as the prior knowledge. We have tried other kernel-based similarity functions to calculate $w_{ip}$, \textit{e.g.}, t-distribution kernel %$w_1 = \frac{(1+ ||\phi_{old}(i)-\phi_{old}(p) ||^2/\alpha )^{-\frac{\alpha+1}{2}}}{\sum_{a\in A(i)} (1+ ||\phi_{old}(i)-\phi_{old}(a) ||^2/\alpha )^{-\frac{\alpha+1}{2}}   } $ ,
$w_{ip}' = \frac{(1+ ||\phi_{old}(i)-\phi_{old}(p) ||^2 )^{-1}}{\sum_{a\in A(i)} (1+ ||\phi_{old}(i)-\phi_{old}(a) ||^2 )^{-1}} $ ,
and gaussian kernel $w_{ip}'' = \frac{\exp(\phi_{old}(i)\phi_{old}(p)/\tau)}{\sum_{a\in A(i)}\exp(\phi_{old}(i)\phi_{old}(a)/\tau)} $. We find that such $w_{ip}$ differs only slightly, and the mean average precision changes smaller than 0.5\% (detailed in supplementary). Therefore, we calculate $w_{ip}$ by Equation (\ref{equation 5}) for simpilfily.

$s_{ip}$ represents the affinity score of positive $p$ contrasted with an anchor $i$, where $\tau$ is a temperature parameter that controls the sharpness of similarity distribution. Minimizing $L_{1}$ is equivalent to maximize the affinity scores between anchors and positives. The existence of weight forces the optimized gradient direction to be more biased towards those positive pairs that are geometrically closer. Therefore, the neighborhood consensus contrastive learning procedure on the embedding space actually ensures the alignment of the old and new embeddings at the granularity of local clusters.

\subsection{Soft multi-label task on the discrimination space}  
  As the embedding space represents the instance-level knowledge, the proposed neighborhood consensus supervised contrastive learning can be regarded as an intra-class alignment between old and new embeddings. While the discrimination space represents the class-level knowledge, which defines the probability distribution of each sample belonging to the underlying class centers on the simplex \cite{zhang2021unleashing}. Intuitively, we can impose constraints on classifiers to achieve the class-level alignment and tightening.

With the new classifier, we construct a pseudo-multi-label task from a different perspective than semantic-level label learning. 
% facet 
Specifically, given image $i$, its positive image set $P$ and negative image set $N$, we compare the discrimination vector $\varphi_{new}(\phi_{new}(i))$ with discrimination vector set $\{\varphi_{new}(\phi_{old}(p)), p \in P\}$ and $\{\varphi_{new}(\phi_{old}(n)), n \in N\}$. Intuitively, $\varphi_{new}(\phi_{new}(i))$ should be similar to any positive discrimination vector $\varphi_{new}(\phi_{old}(p))$ and dissimilar to any negative discrimination vector $\varphi_{new}(\phi_{old}(n))$. Thus, the positive/negative discrimination vectors can be regarded as pseudo-multi-labels as a supplement to ground truth. As the ground truth may only reflect a single facet of the complete knowledge encapsulated in real-world data \cite{xu2020knowledge}, such label is an auxiliary to mine richer knowledge, resulting in the new embeddings with better performance. Here we imitate the contrastive procedure in the embedding space for harmonization. A FIFO memory queue is also utilized to store the positive/negative discrimination vector. We give the dual soft contrastive loss on the discrimination space below, 

  \vspace{-0.3cm} 
\begin{equation}
L_{2}  = \sum_{i\in \mathcal{D}_{new}}\sum_{p\in P(i)} -w_{ip}\log \tilde{s}_{ip},
\end{equation}
  \vspace{-0.4cm} 
\begin{equation}
\tilde{s}_{ip} =  \frac{\exp(\varphi_{new}(\phi_{new}(i))\varphi_{new}(\phi_{old}(p))/\tau)}{\sum_{a\in A(i)}\exp(\varphi_{new}(\phi_{new}(i))\varphi_{new}(\phi_{old}(x_a))/\tau)}. 
\label{equation 7}
\end{equation}
  \vspace{-0.55cm} 
  
\noindent
where the weight $w_{ip}$ is the same as in Equation (\ref{equation 5}).

\textbf{Discussion.}  % 还是有点抽象  具体实验结果/toy 实验结果  比较苍白
Regarding how to constrain the new and old discrimination vectors, there are the following options: 
1) Constrain the new and old discrimination vectors to be consistent with their one-hot semantic label (\textit{e.g.}, using the cross-entropy). Such vectors with the same ID tend to fall in the same hyperplane block. However, it is difficult to converge to an ideal one-hot form, especially in a large amount of IDs scenario. Thus, the distance between new and old embeddings with the same ID may be large, which means backward compatibility is not ensured.
2) Constrain the new discrimination vectors to be consistent with their corresponding old ones (\textit{e.g.}, using the KL divergence). Such vectors corresponding to the same image tend to be consistent. However, the divergence or variance of the discrimination distribution for each class may be larger, impacting the performance of the new model.
3) To achieve compatibility with the new embeddings more compact, we jump out of the constrain of pair-based methods. We use the discrimination vectors of other samples with the same ID in the neighborhoods to constrain each other. As a result, such new vectors tend to be consistent at a cluster level, as a new vector and its neighbor are regularized by the same old discrimination vectors.
%To better compress the variance, we jump out of the comparison of discrimination probabilities in a single sample and use the discrimination vectors of other samples with the same ID in the neighborhoods to constrain each other. Once the discrimination distribution divergence is reduced, the class structure would become compact. 

%If we only use the semantic-label guided cross-entropy loss to optimize the new and old discrimination probabilities of images, it ensures that these two probability vectors from the same sample fall in the same hyperplane block. Nevertheless, their distance on the simplex space cannot be guaranteed to be small since the final convergent probability vectors often do not tend to the ideal one-hot form, especially in a large amount of IDs scenario. Similarly, if we penalize the KL divergence of these two probability vectors, the ideal convergence state is that they are consistent. However, the divergence or variance of the discrimination distribution for each class may still be large. To better compress the variance, we jump out of the comparison of discrimination probabilities in a single sample and use the discrimination vectors of other samples with the same ID in the neighborhoods to constrain each other. Once the discrimination distribution divergence is reduced, the class structure would become compact. 

\subsection{Highly credible sample detection}
Due to the variations of object morphology, illumination, and camera view in the real-world data, the intra-class variance of old embeddings can be huge \cite{bai2018group}. Poor old embeddings are often located in the fuzzy region between classes. Compatible learning with such outlier samples would affect the retrieval accuracy of the new model. 

Here, we introduce entropy to measure the uncertainty of sample classification. With greater entropy, the classification vectors tend to have smaller credibility. One straightforward method to construct the classification vectors is to use the old classifier. However, the old classifier can not handle newly added classes. Besides, it is not available when a third party deploys the old model, as shown in Figure \ref{deploy}. %The reason to give up the classifier is as following: 1) As only embeddings are needed for retrieval, the classifier only works in training progress but is useless for inference; 2) There are usually many classes for training, making the classifier taking up a lot of space. 
Therefore, we construct the pseudo classification vectors $\tilde{p}(x)$ by calculating the similarity score of each sample belonging to old class centers. Specifically, suppose the $\hat{\mu}_{k}=\frac{1}{|\{i: y_{i}=k\}|}\sum_{y_{i}=k}\phi_{old}(i)$ is the $k$-th ($1\leq k\leq K_{new}$) old class center. Here we test the reliability of $\hat{\mu}_{k}$ through a simple experiment: for the model ResNet18 trained by 50\% data on Market1501, the mAP is only 63.26\%. However, the discrimination accuracy of using the nearest geometry center is 96.35\% and 99.17\% for the test set and training set, respectively, proving its effectiveness. Then we utilize multiple Gaussian kernel functions to obtain the similarity score between samples and centers, 
%\begin{align}
%\tilde{p}(x_{i})_{k} & = \frac{\exp(-\frac{1}{2}(\phi_{old}(x_{i})-\hat{\mu}_{k})^{T}\hat{\Sigma}_{k}^{-1}(\phi_{old}(x_{i})-\hat{\mu}_{k}))}{\sum_{j=1}^{K_{new}}\exp(-\%frac{1}{2}(\phi_{old}(x_{i})-\hat{\mu}_{j})^{T}\hat{\Sigma}_{k}^{-1}(\phi_{old}(x_{i})-\hat{\mu}_{j}))}
%\end{align}
  \vspace{-0.3cm} 
\begin{equation}
    \tilde{p}(i)_{k} = \frac{\exp(-\frac{||\phi_{old}(i)-\hat{\mu}_{k}||^{2}}{\hat{\sigma}_{k}})}{\sum_{j=1}^{K_{new}}\exp(-\frac{||\phi_{old}(i)-\hat{\mu}_{j}||^{2}}{\hat{\sigma}_{j}})}.
\end{equation}

  \vspace{-0.3cm} 

Where $\hat{\sigma}_{k}$ represents the variance of $k$-th distance set $\{||\phi_{old}(i)-\hat{\mu}_{k}||^{2}: y_{i}=k\}$. Then the uncertainty of $i$-th sample is the entropy of $\tilde{p}(i)$, namely $\mathcal{H}(i)=\sum_{k=1}^{K_{new}}-\tilde{p}(i)_{k}\log \tilde{p}(i)_{k}$. Since $\log K_{new}$ is the maximum value of $\mathcal{H}$, the uncertainty threshold boundary is set as $\hat{U}$ which is related to $\log K_{new}$. 
Then we remove those samples with entropy larger than $\hat{U}$, and they do not participate in the soft supervised contrastive learning procedure. 

\textbf{Overall loss function.} %Combining with the cross-entropy based classification loss $L_{new}$ on the new model, we give the final objective below,   
Combining with classification loss $Loss_{new}$ on the new model trained independently, we give the final objective below, 
  \vspace{-0.2cm} 
\begin{equation}
L_{total} = L_{new} + \alpha L_{1} + \beta L_{2}.
\end{equation}
  \vspace{-0.5cm}
 
  \noindent
Where the $\alpha$ and $\beta$ are two hyperparameters to control the relative importance of embedding space learning and discrimination space learning, respectively. %Specifically, we first use $L_{new}$ and $L_{1}$ loss to train the entire new model. Then we freeze the weight of the classification head before adding the $L_{2}$ loss in training to avoid drastic changes in the old discrimination features. This step also actually helps stabilize the training of pseudo-multi-label learning.

\begin{table*}[t]
  \centering
  %\small
  \fontsize{7.5}{7}\selectfont
     \setlength{\tabcolsep}{3.3mm}
  \caption{Performance comparison on Market1501, MSMT17 and VeRi-776 datasets. }
      %\vspace{-0.6cm} 
  \renewcommand\arraystretch{1.2}
  \label{baseline}
  \begin{center}
    \begin{tabular}{l|cccc|cccc|cccc}
      \hline
      \multicolumn{1}{c}{~} & 
      \multicolumn{4}{|c}{\textbf{Market1501}}         & \multicolumn{4}{|c}{\textbf{MSMT17}} & \multicolumn{4}{|c}{\textbf{VeRi-776}} \cr   
      \cline{2-13}
      \multicolumn{1}{c|}{\textbf{Methods}} &
      \multicolumn{2}{c}{\textbf{Self-Test}}         & \multicolumn{2}{|c}{\textbf{Cross-Test}}   &
       \multicolumn{2}{|c}{\textbf{Self-Test}}         & \multicolumn{2}{|c}{\textbf{Cross-Test}}   &
       \multicolumn{2}{|c}{\textbf{Self-Test}}         & \multicolumn{2}{|c}{\textbf{Cross-Test}}   
      \cr
      
      \cline{2-13}
      
      \multicolumn{1}{c|}{~}   & 
      \multicolumn{1}{c}{R1} &   \multicolumn{1}{c|}{mAP} & 
      \multicolumn{1}{c}{R1} &  \multicolumn{1}{c|}{mAP} &
      \multicolumn{1}{c}{R1} &  \multicolumn{1}{c|}{mAP} &
      \multicolumn{1}{c}{R1} &  \multicolumn{1}{c|}{mAP} &
      \multicolumn{1}{c}{R1} &  \multicolumn{1}{c|}{mAP} &
      \multicolumn{1}{c}{R1} &  \multicolumn{1}{c}{mAP} \cr \hline % \hline
      Ori-Old(R18) & 82.96 & 63.26 &-  &-  & 57.02 & 29.24 &-  &-  & 89.35 & 56.91 &-  &-\cr   \hline

      $L_2$ regression   &91.66 & 78.69 & 3.03 & 2.27 & 67.3 & 39.57 & 0.51 & 0.26 & 91.13 & 63.75 & 4.23 & 2.97
      \cr
      BCT  &91.69 & 77.47 & 84.74 & 66.86 & 67.91 & 39.11 & 58.18 & 30.44 & 92.02 & 64 & 80.86 & 56.41
      \cr
      Asymmetric &91.98 & 81.08 & 85.45 & 68.4 & 71.95 & 45.29 & 53.90 & 28.15 & 91.26 & 66.59 & 81.01 & 56.28
      \cr
      NCCL(Ours) &\textbf{92.87} & \textbf{81.90} & \textbf{85.51} & \textbf{69.15} & \textbf{73.43} & \textbf{46.00} & \textbf{59.64} & \textbf{30.92} & \textbf{92.62} & \textbf{70.15} & \textbf{81.37} & \textbf{58.82}

      \cr
      \hline
      Ori-New(R18)  &92.37 & 80.84 & 22.71 & 13.81 & 70.52 & 42.86 & 10.06 & 4.11 & 92.26 & 67.21 & 19.70 & 10.00\cr   \hline \hline 
      Ori-Old(R50)  &87.23 & 70.48 & - &-  & 64.29 & 36.45 & - & - & 91.90 & 62.91 &-  &- 
\cr   \hline

      $L_2$ regression  & 93.97 & 84.31 & 47.30 & 28.97 & 72.43 & 47.16 & 30.91 & 13.22 & 93.15 & 70.61 & 47.50 & 25.64

      \cr
      BCT  & 93.35 & 82.51 & 90.50 & 75.21 & 71.25 & 45.07 & 68.77 & 40.45 & 92.02 & 67.49 & 88.44 & 64.89

      \cr
      Asymmetric & 94.39 & 86.42 & 90.32 & 77.34 & 75.37 & 51.44 & 70.43 & 41.68 & 92.27 & 70.19 & 89.70 & 63.41
%55.97 80.55
      \cr
      NCCL(Ours) & \textbf{94.80} & \textbf{87.24} & \textbf{91.60} & \textbf{77.69} & \textbf{77.55} & \textbf{52.97} & \textbf{71.45} & \textbf{42.87} & \textbf{94.4} & \textbf{75.42} & \textbf{89.70} & \textbf{66.74}

      \cr
      \hline
      Ori-New(R50) & 94.30 & 85.10 & 81.98 & 60.78 & 73.58 & 48.42 & 52.46 & 26.47 & 93.69 & 70.70 & 65.48 & 37.90\cr   \hline \hline
      
    \end{tabular}
    %\vspace{-5pt}
  \end{center}
      \vspace{-0.5cm} 
\end{table*}
\section{Experiments}
We evaluate our proposed method on two widely-used person ReID datasets, \textit{i.e.}, Market-1501~\cite{zheng2015scalable}, MSMT17~\cite{wei2018person}, and one vehicle ReID dataset VeRi-776~\cite{wang2017orientation}.
We first implement several baselines, and then test the potential of our method by applying it to the case of multi-factor changes, including model changes and loss changes. We also conduct a multi-model test to evaluate the sequential compatibility.
%A multi-model test also evaluates the sequential compatibility. 
Besides, we adopt the large-scale ImageNet~\cite{deng2009imagenet} and Place365~\cite{zhou2017places} datasets to demonstrate the scalability of our method.

\subsection{Datasets and evaluation metrics}
Market1501 consists of 32,668 annotated images of 1,501 identities shot from 6 cameras in total. MSMT17 is a large-scale ReID dataset consisting of 126,441 bounding boxes of 4,101 identities taken by 15 cameras. VeRi-776 contains over 50,000 images of 776 vehicles captured by 20 cameras.% covering a 1.0 $km^2$ area in 24 hours. 
ImageNet contains more than 1.2 million images with 1,000 classes. Place365 has about 1.8 million images with 365 classes. For ImageNet and Place365, we conduct the retrieval process on the validation sets of these two datasets, i.e., each image will be taken as a query image, and other images will be regarded as gallery images.

Mean average precision (mAP) and top-$k$ accuracy are adopted to evaluate the performances of retrieval. $M(\phi_{new}, \phi_{new})$ (termed as ``self-test'') and $M(\phi_{new}, \phi_{old})$ (termed as ``cross-test'') are used to evaluate the performances of backward-compatible representation.%The self-test performance and cross-test performance are used to evaluate the performances of backward-compatible representation. Self-test means that $\phi_{new}$ is used to extract embeddings for both the query set and the gallery set. Cross-test means that $\phi_{new}$ is used to extract embeddings for the query set while $\phi_{old}$ is used to compute embeddings for the gallery set. 

\subsection{Implementation details} %待重写
The implementations of all compared methods and our NCCL are based on the FastReID~\cite{he2020fastreid}. The default configuration in ``SBS'' is adopted for the model using ``CircleSoftmax'' and ``ArcSoftmax''. And the default configuration in ``bagtricks'' is adopted for the model using ``Softmax'' with slight changes: size\_train/size\_test is set to 384x128 to be consistent with ``SBS''. The size of  $\mathcal{B}$ is 2048 and $\tau$ is 1.0. We set $\alpha$ and $\beta$ around 0.01 so that $L_1$ loss and $L_2$ loss are on the same order of magnitude as $L_{new}$ loss. $\hat{U}$ is set to $\frac{\log K_{new}}{2}$ as the credible sample selection threshold. More detail is shown in the suplementary.

%The default configuration in ``bagtricks'' is adopted for the model use ``softmax'', and that of ``SBS'' is adopted for the model use ``circlesoftmax'' and ``arcsoftmax'', with slight changes: the resolution is set to $384x128$ 
%We use 4 NVIDIA Tesla V100 GPUs for training. The input size of the models is set to $384 \times 128$ pixels. The implementations of all compared methods, including $L_{2}$ regression, BCT, Asymmetric, and our NCCL, are based on the FastReID~\cite{he2020fastreid}. The default configuration of FastReID is adopted, including weight decay, learning rate, etc. The parameters $\alpha$, $\beta$, and $\tau$ are taken as 0.01, 0.01, 1.0 by default, respectively. We set $\hat{U}=\frac{\log K_{new}}{2}$ as the default credible sample selection threshold. Without additional explanation, we use ResNet-18 \cite{he2016deep} as the backbone to obtain a 512-dimensional embedding feature vector and cross-entropy as $L_{new}$. In our NCCL, the soft pseudo-multi-label learning only works in the later training stage, with the classifier froze. 

\subsection{Baseline comparisons}
In this section, we conduct several baseline approaches with backbone ResNet-18 and ResNet-50. With the same model architectures and loss functions, we set the old training data of 50\% IDs subset and the new training data of 100\% IDs complete set. The results are shown in Table \ref{baseline}. %The cross-tests and self-tests verified the effectiveness of the proposed method. Besides, surprisingly, our proposed method outperforms the new model trained without any regularization.

\textbf{No regularization between $\phi_{new}$  and $\phi_{old}$. } We denote the new model trained without any regularization as $\phi_{new*}$. A simple case directly performs a cross-test between $\phi_{new*}$ and $\phi_{old}$. As is shown in Table \ref{baseline}, backward compatibility is achieved at a certain degree. 
However, the cross-test accuracy fluctuates dramatically with the change of data set and network architecture, which is insufficient to satisfy our backward-compatibility criterion. Subsequent experimental results will further confirm this opinion.

\textbf{$L_2$-regression between $\phi_{new}$ and $\phi_{old}$ output embeddings.} An intuitive idea to achieve backward-compatibility is using $L_2$ loss to minimize the output embeddings' euclidean distance between $\phi_{new}$ and $\phi_{old}$, which was discussed in \cite{Shen2020}. 
Such a simple baseline could not meet the requirement of backward-compatibility learning. The possible reason for $L_2$ loss failure is that $L_2$ loss is too local and restrictive. It only focuses on decreasing the distance between feature pairs from the same image, ignoring the distance restriction between negative pairs.

\textbf{Asymmetric.} 
It works better on small datasets than on large-scale datasets, probably owing to its inadequacy in utilizing the intrinsic data structure. For instance,  Asymmetric's cross-test outperforms the old model's self-test on Market1501, but not on VeRi-776, as is shown in Table \ref{baseline}.

\textbf{BCT.}  
As is shown in Table \ref{baseline}, BCT underperforms $\phi_{new*}$ in all self-test cases. The most likely reason is that BCT use synthesized classifier weights or knowledge distillation to deal with the newly added data. Synthesized classifier weights for the newly added data are susceptible to those boundary points or outliers, making the training results unpredictable. Knowledge distillation assumes that the teacher model (old model) outperforms the student model (new model), which is not true in the application scenario of backward-compatible representation. As a result, it impairs the performance of the new model.

\textbf{NCCL(Ours) with better self-test accuracy.} 
As the effective utilization of the intrinsic structure, NCCL outperforms all compared methods. Besides, it significantly outperforms $\phi_{new*}$ in all self-test cases. Although all the data are accessible from the new model, the old model provides external knowledge due to its different architectures, initializations, training details, etc \cite{zhang2018deep}. The key to better self-test accuracy is in extracting useful knowledge from the old model and eliminating the impact of outliers in the old embeddings. BCT does not take it into account but distills at the instance level and constrains the distributions of the new embeddings to be consistent with the old. We use entropy to measure the uncertainty of sample classification, so most outliers are filtered. Besides, we take a contrastive loss with multiple positives and negatives at the sub-class level. Thus the effects of the outliers are diminished as multiple old embeddings serve as ``mean teachers'', similar to \cite{2020Improving}. Such loss constrains the distance relationship between the new and old embeddings to be consistent.
%With the proposed method, the new embeddings could be improved with the old model's information. 

\subsection{Changes in network architectures and loss functions}
Here we study whether each method can be stably applied to various model changes and loss function changes. 

 \begin{table}[t]
\caption{Performance comparison between different models on Market1501. The old model uses ResNet-18 with 50\% training data. ``mAP1'' and ``mAP2'' represent the mean average precision of self-test and cross-test, respectively. }
      %\vspace{-0.4cm} 

\label{modelchanges}
    \setlength{\tabcolsep}{2.5mm}
\centering
  \fontsize{6.7}{7.5}\selectfont
\begin{tabular}{l|cc|cc|cc} \hline
    \multicolumn{1}{c|}{~} & \multicolumn{2}{c|}{\textbf{ResNet18-Ibn}} & \multicolumn{2}{c|}{\textbf{ResNet50}} & \multicolumn{2}{c}{\textbf{ResNeSt50}} \\ \cline{2-7}
    \multicolumn{1}{c|}{\textbf{Methods}} & mAP1 & mAP2 & mAP1 & mAP2& mAP1 & mAP2 \\ \hline 
Ori-Old    &63.26  & - &63.36 &- &63.26    & -      \\\hline
$L_2$ regression              & 80.05 &  1.19   &84.54  & 0.56& 83.03& 0.59 \\
BCT              &78.92 &  66.70  &82.03  & 68.65& 86.64& 68.76\\
Asymmetric       &82.23 &  68.54  &83.96  & 67.31 &87.81 & 68.76 \\
NCCL(Ours)   &\textbf{83.23}  &  \textbf{69.12}    &\textbf{85.91}  & \textbf{69.26} & \textbf{88.31}& \textbf{68.94}     \\  \hline
Ori-New &81.59 & 5.12   &84.88  & 0.20& 88.21 & 0.20 \\\hline
\end{tabular}
     % \vspace{-0.4cm} 

\end{table}
\textbf{Model Changes.} We first test the new model on ResNet18-Ibn, ResNet-50 and ResNeSt-50 \cite{zhang2020resnest} instead of the old model ResNet-18 under the same loss function. ResNet18-Ibn only changes the normalization method of the backbone from batch normalization to instance batch normalization \cite{pan2018two}, and it keeps the structure consistent. It can be seen as a slight modification to the old model. ResNet-50 is composed of the same components with different capacities 
compared to ResNet-18, 
which can be regarded as a middle modification to the old model. ResNeSt-50 introduces the split attention module into ResNet50, which makes it the most different from ResNet18. Note that the dimension of the feature vector in ResNet-50 and ResNeSt-50 is 2048. We will not directly feed the old feature with dimension 512 to $\phi_{new}$ but use the zero-padding method to expand it to 2048-dimension.

As shown in Table \ref{modelchanges}, the comparison between independently trained new models and old models is an epic failure. BCT, Asymmetric, and our proposed method have learned compatible representation. Besides, our approach remains the the-state-of-art and is more robust to model changes.

\begin{table}[t]
\caption{Performance comparison between different losses on Market1501. The old model uses ResNet-18 with 50\% training data. ``mAP1'' and ``mAP2'' represent the mean average precision of self-test and cross-test, respectively.}
     % \vspace{-0.4cm} 
\label{LossChanges}
    \setlength{\tabcolsep}{2.5mm}
\centering
  \fontsize{6.7}{7.5}\selectfont
\begin{tabular}{l|cc|cc|cc} \hline
    \multicolumn{1}{c|}{~} & \multicolumn{2}{c|}{\textbf{CircleSoftmax}} & \multicolumn{2}{c|}{\textbf{ ArcSoftmax} }& \multicolumn{2}{c}{ \textbf{Softmax+Triplet} }\\ \cline{2-7}
    \multicolumn{1}{c|}{\textbf{Methods}} & mAP1 & mAP2 & mAP1 & mAP2 & mAP1 & mAP2 \\ \hline 
Ori-Old              &63.26  & - &63.36 &- &63.26  &-        \\\hline
$L_2$ regression              & 62.67 &  3.29   &54.03  & 0.85 & 81.27& 23.53 \\
BCT              &67.63 &  \textbf{53.20}  &63.49  & \textbf{39.96} & 79.62 & 67.43\\
Asymmetric       &67.27 &  28.77  &56.71  & 5.09 & 81.75 & 68.07 \\
NCCL(Ours)   &\textbf{79.62}  &  49.28    &\textbf{71.06}  & 31.38 & \textbf{82.36}& \textbf{69.15}     \\  \hline
Ori-New &78.96 & 3.69   &67.31  & 0.93& 81.55 & 14.20\\ \hline
\end{tabular}
     % \vspace{-0.4cm} 
\end{table}

\textbf{Loss Changes.} Then we change the “Softmax” based loss function $L_{new}$ to ``ArcSoftmax'' \cite{deng2019arcface}, ``CircleSoftmax'' \cite{sun2020circle} and ``Softmax + Triplet'' in the new model. The first two are both used in the classification head, and the last “Triplet” is used in the embedding head.

As shown in Table \ref{LossChanges}, the performances of various methods are consistent with the baseline under the change of embedding loss. In contrast, all methods have a significant decline in self-test and cross-test under classification loss changes. 
The possible reason is that classification loss in the new models determines the class structure, while the embedding loss tightens the sample's representation. 
Among these methods, Asymmetric fails in both self-test and cross-test. BCT achieves slightly better performance in cross-test but significantly reduces new models' performance, making it unnecessary to deploy new models. Our method can improve the performance of the new model stably while ensuring the performance of the cross-test.

\subsection{No overlap between $\mathcal{D}_{old}$ and $\mathcal{D}_{new}$}
When the old model is from a third party, all of its information, including the classifier and training sets, are unavailable, except for the embedding backbone. 
In the previous experiments, we always assume that the $\mathcal{D}_{new}$ is a superset of $\mathcal{D}_{old}$. Here we investigate whether no overlap between them could sharply affect the compatibility performance of each method. The old model is trained with the randomly sampled 25\% IDs subset of training data in the Market1501 dataset, and the new model is trained with the other 75\% IDs subset. We show the results in Table \ref{overlap}.

Compared with baseline, the accuracy of all evaluated methods has dropped, except for NCCL. The performance degradation of BCT is the most severe because it can only use synthesized classifier weights or knowledge utilization in this case. Both of them can be regarded as a kind of knowledge distillation at the instant level. As the old model is not good enough, BCT is difficult to extract useful knowledge and eliminate the impact of outliers in the old embeddings. Our model has carefully considered such cases. Thus we achieve the same effect as the baseline in this case. 
%It is intuitive to use old models that are not good enough to distill knowledge and obtain unsatisfactory results. 

%Our method does not require the old classifier, and the processing method for overlapping and non-overlapping training data is consistent. As we use sample selection and neighbor relationship maintenance, our method achieves the same effect as the baseline in this case. 

\begin{table}[t]
\caption{Performance comparison of no overlap between $\mathcal{D}_{old}$ (25\% IDs) and $\mathcal{D}_{new}$ (75\% IDs) on Market1501. }
     % \vspace{-0.4cm} 
\label{overlap}
\centering
 \fontsize{6.7}{7.5}\selectfont
   \setlength{\tabcolsep}{2.8mm}
\begin{tabular}{l|ccc|ccc}
  \hline
  %\multicolumn{1}{c}{\multirow{~}} & \multicolumn{6}{|c}{\textbf{Market1501}} \cr   \cline{2-7}
  \multicolumn{1}{c|}{~} &
  \multicolumn{3}{c}{\textbf{Self-Test}}         & \multicolumn{3}{|c}{\textbf{Cross-Test}}  
  \cr
  
  \cline{2-7}
  
  \multicolumn{1}{c|}{\textbf{Methods}}   & R1 & R5 &mAP & R1 &R5 &mAP \cr \hline
  Ori-Old & 70.13 & 86.07 & 46.42 &  - & -  & -\cr   \hline

  $L_2$ regression  & 90.29 & 96.32 & 76.24 & 15.86 & 35.33 & 9.41 \cr
  BCT  & 86.73 & 94.86 & 69.60 & 62.29 & 83.49 & 41.63 \cr
  Asymmetric & 91.03 & 96.67 & 78.91 & 74.55 & 90.23 & 53.07 \cr
  NCCL(Ours) & \textbf{92.22} & \textbf{97.09} & \textbf{80.88} & \textbf{77.73} & \textbf{90.94} & \textbf{55.91} \cr
  \hline
  Ori-New & 90.17 & 96.32 & 77.52 &  6.00 & 16.60 & 3.92\cr   \hline

\end{tabular}
     % \vspace{-0.3cm} 
\end{table}

\subsection{Multi-model and sequential compatibility}
%Here we investigate the sequential compatibility case of multi-model. The first model $\phi_{1}$ is trained with the randomly sampled 25\% IDs subset of training data in the Market1501 dataset, and the second model $\phi_{2}$ is trained with a 50\% subset. Finally, the third model $\phi_{3}$ is trained with the full dataset. We train $\phi_{2}$ using compatible methods with $\phi_{1}$ and train $\phi_{3}$ using compatible methods with $\phi_{2}$. Therefore, $\phi_{3}$ has no direct influence on $\phi_{1}$. The results of the sequential compatibility test are shown in Table \ref{seqtest}. 
Here we investigate the sequential compatibility case of multi-model. There are three models: $\phi_{1}$, $\phi_{2}$ and $\phi_{3}$. $\phi_{1}$ is trained with the randomly sampled 25\% IDs subset of training data in the Market1501 dataset. $\phi_{2}$ is trained with a 50\% subset. And $\phi_{3}$ is trained with the full dataset. We constrain $\phi_{2}$ to be compatible with $\phi_{1}$ and $\phi_{3}$ to be compatible with $\phi_{2}$. Thus, $\phi_{3}$ has no direct influence on $\phi_{1}$. The results of the sequential compatibility test are shown in Table \ref{seqtest}. 

We observe that, by training with NCCL, the last model $\phi_{3}$ is transitively compatible with $\phi_{1}$ even though $\phi_{1}$ is not directly involved in training $\phi_{3}$. It shows that transitive compatibility between multiple models is achievable through our method, enabling a sequential update of models. It is worth mentioning that our approach still has significant advantages over other methods in cross-model comparison $\phi_{3}2\phi_{1}$.

\begin{table}[t]
\caption{Performance comparison in sequential multi-modal experiment on Market1501. We train three models $\phi_{1}$ (ResNet18+25\% IDs), $\phi_{2}$ (ResNet18+50\% IDs) and $\phi_{3}$ (ResNet18+100\% IDs).}
      %\vspace{-0.4cm} 
\label{seqtest}
\centering
  \fontsize{6.7}{7.5}\selectfont
  \label{tab5}
  \centering
  \setlength{\tabcolsep}{0.9mm}
\begin{tabular}{l|cc|cc|cc} \hline
    \multicolumn{1}{c|}{~} & \multicolumn{6}{c}{ \textbf{Market1501 (mAP)}}  \cr \cline{2-7}
    \multicolumn{1}{c|}{ \textbf{Methods} } & $(\phi_{1},\phi_{1})$ & $(\phi_{2},\phi_{1})$ & $(\phi_{2},\phi_{2})$ & $(\phi_{3},\phi_{2})$ & $(\phi_{3},\phi_{3})$ & $(\phi_{3},\phi_{1})$ \\ \hline 
Ori  & 46.42 & 7.49 & 63.30  & 14.15 & 80.85  & 5.62\\ \hline
$L_2$ regression   & - & 24.01 & 61.87  & 16.15 & 78.39  & 1.47 \\
BCT  & - & 45.78 & 60.29  & 65.45 & 77.07 & 45.20\\
Asymmetric &- & 45.68 &  62.88  & 67.51 & 81.28 & 48.85 \\
NCCL(Ours)   & - & \textbf{45.89} & \textbf{64.27}  & \textbf{69.76} & \textbf{82.28} & \textbf{52.76} \\
\hline
\end{tabular}
     % \vspace{-0.2cm} 
\end{table}

\subsection{Test on other two large-scale image datasets}
As the need for backward-compatible learning stems from the fact that the gallery is too large to re-extract new embeddings, we also utilize large-scale ImageNet and Place365 datasets for training and evaluating each method. The results are shown in Table \ref{tab6}. On large-scale datasets, the accuracy of Asymmetric drops significantly. NCCL achieves better performance than baseline. Asymmetric is too local and restrictive to capture the intrinsic structure in the large datasets, while our method has a solid ability to obtain the intrinsic structure.

\begin{table}[t]
  \centering
  %\small
  \fontsize{7.5}{7}\selectfont
    \setlength{\tabcolsep}{1.1mm}
  \caption{Performance comparison on two large-scale datasets, ImageNet and Place365. The new model uses ResNet-18 with 50\% Data, while the new model uses ResNet-50 with 100\% Data.}
  \vspace{-0.5cm}
  \renewcommand\arraystretch{1.3}
  \label{tab6}
  \begin{center}
    \begin{tabular}{l|cccc|cccc}
      \hline
      \multicolumn{1}{c}{~} & 
      \multicolumn{4}{|c}{\textbf{Place365}}         & \multicolumn{4}{|c}{\textbf{ImageNet}} \cr   
      \cline{2-9}
      \multicolumn{1}{c|}{\textbf{Methods}} &
      \multicolumn{2}{c}{\textbf{Self-Test}}         & \multicolumn{2}{|c}{\textbf{Cross-Test}}   &
       \multicolumn{2}{|c}{\textbf{Self-Test}}         & \multicolumn{2}{|c}{\textbf{Cross-Test}} \cr
      
      \cline{2-9}
      
      \multicolumn{1}{c|}{~}   & 
      \multicolumn{1}{c}{Top-1} &  \multicolumn{1}{c|}{Top-5} & 
      \multicolumn{1}{c}{Top-1} &  \multicolumn{1}{c|}{Top-5} &
      \multicolumn{1}{c}{Top-1} &  \multicolumn{1}{c|}{Top-5} &
      \multicolumn{1}{c}{Top-1} &  \multicolumn{1}{c}{Top-5} 
      \cr 
      \hline
      Ori-Old  & 27.0 & 55.9 & - & -  & 39.5 &60.0  & - & - \cr   
      \hline
      BCT  & 32.7 & 62.2 & 27.6 & 57.6   &41.9 &65.3  &42.2 &65.5 \cr
      Asymmetric & 31.2 & 60.9 & 28.6 & 58.1   &41.8 &65.1  & 42.9 & 63.3 \cr
      NCCL(Ours) & \textbf{35.1} & \textbf{64.1} & \textbf{28.7} & \textbf{58.1} &\textbf{62.5} &\textbf{81.6}  & \textbf{46.1} & \textbf{68.2} \cr
      \hline
      Ori-New   &35.1 & 64.0 & - & - &62.5 &81.5  &- & -\cr  
      \hline
    \end{tabular}
   % \vspace{-5pt}
     \vspace{-0.3cm}
  \end{center}
\end{table}

\subsection{Discussion}
Generally, three motivations drive us to deploy a new model: 1) growth in training data, 2) better model architecture, and 3) better loss functions. We have conducted detailed experiments on these three aspects. BCT can deal with the data growth and model architecture changes when the old and new training data have common classes. Asymmetric can handle data growth and model architecture changes well on small datasets, even better than BCT, but it is too local to work on large datasets reliably. Our method is state-of-the-art under the changes of training data and model architectures. It achieves better cross-test performance and gives better performance than the new model trained independently. 

For loss functions, the change of embedding loss usually does not affect backward-compatible learning. In contrast, the change of classification loss will lead to both BCT failure and Asymmetric failure, which means that the cross-test performance drops sharply or the new model's performance is significantly damaged. However, our method can still improve the performance of the new model under the variation of classification loss. At the same time, the backward-compatible version is obtained.
%这一段或许可以放前面改变loss的时候？

\subsection{Ablation study}
We test the effect of different modules in NCCL. 
As is shown in Table \ref{Ablation}, we can observe that adaptive neighborhood consensus helps to strengthen both self-test and cross-test performance significantly. The soft pseudo-multi-label learning also improves performance steadily but not as significantly as the neighborhood consensus module.
Besides, the proposed highly credible sample detection procedure does not bring significant performance up-gradation in all cases, which depends on the quality of the old model. 

%We also vary the three primary hyperparameters: $\alpha$, $\beta$ and $\hat{U}$. The $\alpha$ and $\beta$ control the importance of the $L_{1}$ and $L_{2}$, while $\hat{U}$ is the uncertainty threshold for highly credible sample detection. For simplicity, we take $\alpha = \beta$ and change them from 0.005 to 0.015 and $\hat{U}$ from $\frac{\log K_{new}}{5}$ to $\log K_{new}$. As is shown in Figure \ref{fig:parms}, our method is robust to these hyperparameters changes.

\begin{table}[t]
\caption{Ablation study on Market1501. The new model is trained with 100 \% Data using ResNet-18.}
      %\vspace{-0.4cm} 

\label{Ablation}
  \fontsize{6.5}{7.5}\selectfont
     \setlength{\tabcolsep}{1.2mm}
\begin{tabular}{l|ccc|ccc}
  %\hline
  %\multicolumn{1}{c}{~} & 
  %\multicolumn{6}{|c|}{\textbf{Market1501}} \cr   
  \cline{1-7}
  \multicolumn{1}{c|}{~} &
  \multicolumn{3}{c|}{\textbf{Self-Test}}         & \multicolumn{3}{c}{\textbf{Cross-Test}}  
  \cr
  
  \cline{2-7}
  \multicolumn{1}{c|}{\textbf{Methods}}   & 
  \multicolumn{1}{c}{R1} & \multicolumn{1}{c}{R5} & \multicolumn{1}{c|}{mAP} & 
  \multicolumn{1}{c}{R1} & \multicolumn{1}{c}{R5} & \multicolumn{1}{c}{mAP}  \cr \hline
  Ori-Old(R18+50 \%Data) & 82.96 & 92.32 & 63.26 & - & - & -\cr \hline
  NCCL w/o neighborhood consensus  & 91.75 & 97.14 & 78.71 & 85.45 & 93.48 & 67.90 \cr
  NCCL w/o pseudo-multi-label learning   & 91.98  &97.48 &80.08  & 85.51 &94.83 & 68.40 \cr
  NCCL w/o sample selection & 92.58  &97.30 & 80.18 & 85.05 &95.07 & 68.72 \cr
  NCCL & 92.87 & 97.58 & 81.9 & 86.05 & 95.34 & 69.15 \cr     \hline \hline
  Ori-Old(R18+25 \%Data) & 70.13 & 86.07 & 46.42 & - & - & - \cr \hline
  NCCL w/o neighborhood consensus  & 91.81 & 97.09 & 79.40 & 76.63 & 91.57 & 55.60  \cr
  NCCL w/o pseudo-multi-label learning  & 91.92 & 97.06 & 80.57 & 78.06 &91.69 & 56.53\cr
  NCCL w/o sample selection & 91.86 &96.97 & 80.13 & 77.35 &91.51 & 55.71 \cr
  NCCL  & 92.64 &97.24 & 81.41 & 78.06 &91.98 & 57.44
  \cr    
  \hline
\end{tabular}
   %   \vspace{-0.3cm} 

\end{table}

    %  \vspace{0.4cm} 
\section{Conclusion}
This paper proposes a neighborhood consensus contrastive learning framework with credible sample selection for feature compatible learning. We validate the effectiveness of our method by conducting thorough experiments in various reality scenarios using three ReID datasets and two large-scale retrieval datasets. 
Our method achieves the state-of-the-art performances and shows convincing robustness in different case studies. 

{\small
\bibliographystyle{ieee_fullname}
\bibliography{egbib}
}

\end{document}